\documentclass[11pt]{article}

\pdfoutput=1

\usepackage[preprint]{acl}

\usepackage{times}
\usepackage{latexsym}
\usepackage{enumitem}

\usepackage[T1]{fontenc}

\usepackage[utf8]{inputenc}

\usepackage{microtype}

\usepackage{inconsolata}
 
\usepackage{graphicx}
\graphicspath{{./figures/}}

\definecolor{Orange}{RGB}{255,140,0}

\title{When More Words Say Less: \\ Decoupling Length and Specificity in Image Description Evaluation}

\author{Rhea Kapur \\ 
  Stanford University \\
  \texttt{rheak@stanford.edu} \\\And
  Robert Hawkins \\
  Stanford University \\
  \texttt{rdhawkins@stanford.edu} \\\And
  Elisa Kreiss \\
  University of California, Los Angeles \\
  \texttt{ekreiss@ucla.edu} \\}

\begin{document}
\maketitle
\begin{abstract}
Vision-language models (VLMs) are increasingly used to make visual content accessible via text-based descriptions. 
In current systems, however, description specificity is often conflated with their length. 
We argue that these two concepts must be disentangled: descriptions can be concise yet dense with information, or lengthy yet vacuous. 
We define specificity relative to a contrast set, where a description is more specific to the extent that it picks out the target image better than other possible images. 
We construct a dataset that controls for length while varying information content, and validate that people reliably prefer more specific descriptions regardless of length. 
We find that controlling for length alone cannot account for differences in specificity; it matters \textit{how} the length budget is applied. 
These results support evaluation approaches that directly prioritize specificity over verbosity.

\end{abstract}

\section{Introduction}
\label{sec:intro}

\begin{table*}[t]
\centering
\begin{minipage}[c]{0.2\textwidth}
  \centering
  \includegraphics[width=0.81\linewidth]{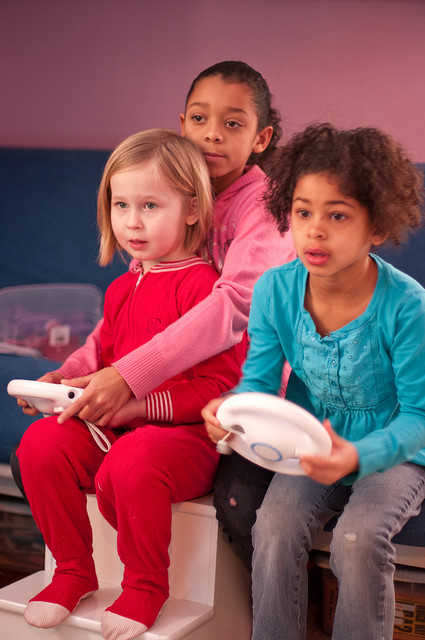}
\end{minipage}\hfill
\begin{minipage}[c]{0.8\textwidth}
  \small
  \centering
  \begin{tabular}{|p{1.7cm}|p{10cm}|}
    \hline
    \textbf{Source} & \textbf{Description} \\
    \hline
    \textsc{original} & There are three girls playing a video game together. \\
    \hline
    \textsc{composite} & Three young girls are sitting next to each other, playing video games together, specifically using Nintendo Wii with wheels. \\
    \hline
    \textsc{verbose} & In the current situation, there are a total of three girls who are engaged in the activity of playing a video game together. \\
    \hline
    \textsc{image-to-text} & The image shows three girls sitting together on a white stool. The girl on the left is wearing a red onesie, the middle girl is dressed in a pink top, and the girl on the right is wearing a blue top. Each girl is holding a game controller. The background features a blue wall and appears to be a living space, likely a playroom or family room. The lighting in the image is warm and soft. \\
    \hline
  \end{tabular}
\end{minipage}
\caption{Example set of descriptions for an image in our dataset. Composite and verbose descriptions are longer variants of the original description, but vary in the amount of additional information provided. Image-to-Text is an example output of a VLM (here, GPT-4o-mini) with minimal instructions. See \autoref{descriptions} for additional examples.}
\label{tab:example_captions}
\end{table*}

Vision–language models (VLMs) are increasingly deployed to produce textual descriptions of visual content \cite{zhang2024vision,wang2024qwen2,deitke2024molmo,ghandi2023deep}, with consequences for blind, low-vision, and sighted users alike \cite{10.1145/2858036.2858116,10.1145/3308558.3313605,stangl2020person}.
When generating descriptions, a central challenge is deciding how \emph{specific} to be: which pieces of information should be included, and at what level of detail?
Whatever determines the appropriate level of specificity for a given context \cite[cf.][]{grice1975logic}, missing that target causes problems: underspecific descriptions fail to support necessary distinctions, while overspecific ones reduce communicative efficiency \cite{goodman2016pragmatic} and can trigger unintended inferences \cite[e.g.,][]{sedivy2003pragmatic,tourtouri2019rational}. 

Specificity is central to communicative effectiveness yet notoriously difficult to measure. 
A natural intuition, grounded in information theory, is that (all else being equal), longer descriptions pack more detailed information \cite{shannon1948mathematical}. 
This intuition motivates a common practice of treating description length as a proxy for specificity, conditional on the description being accurate and well-formed.
This practice appears across studies of human description preferences \cite{williams2022toward,kreiss-etal-2022-context}, dataset construction \cite{urbanek2024picture,wang2025harnessing}, accessibility guidelines \cite{mccall2022rethinking}, and evaluation methods \cite{kapur-kreiss-2024-reference}. 

Yet the relationship between length and specificity is far from straightforward \cite{chen2022grow}: descriptions can be lengthy yet vacuous, or concise yet dense with details.
In fact, we often want to improve a system's specificity and ability to produce distinct outputs while controlling for excessive verbosity  \cite{singhallong, dubois2024length,nayab2024concise,hu2024explaining}. 
For principled evaluation, we must operationalize specificity as independent of length.
Following classic possible world semantics \cite{carnap1947meaning,kripke1959completeness,montague1970universal}, we suggest treating an utterance as more specific when it is compatible with fewer possible worlds. 
In visually grounded settings, a description is more specific when it truthfully describes fewer possible images \cite{young-etal-2014-image,nie-etal-2020-pragmatic}. 
This is an entailment relationship that holds across contexts: ``small red chair'' is strictly more specific than ``red chair''.

In this paper, we construct a dataset that manipulates length independently of information content, pairing images with descriptions that are lengthy yet vacuous (verbose) or concise yet information-dense (composite).
We operationalize specificity via contrastive image compatibility: a description is more specific to the extent that it picks out the target image from a large set of alternatives.
Using this framework, we show that human preferences track specificity, not length, and characterize how different specificity constraints in the prompt (e.g., requesting conciseness versus imposing a character limit) affect the specificity of VLM-generated descriptions. 
Our central finding is that controlling for length alone cannot account for differences in specificity; it matters how the length budget is applied. 
These results support evaluation approaches that directly measure specificity rather than relying on length as a proxy.

\section{Related Work}

\paragraph{Defining specificity via contrast sets}

Recent referring expression generation (REG) models formalize specificity with respect to a contrast set of alternatives \cite{krahmer2012computational}.
In the Rational Speech Act (RSA) framework \cite{goodman2016pragmatic,degen2019redundancyusefulbayesianapproach,degen2023rational}, speakers select utterances that maximize the likelihood of a listener identifying the intended referent from these alternatives while minimizing production costs. 
A key insight from this work is that not all words contribute equally to specificity: it is the inclusion of distinguishing features that differentiate the target from alternatives, not sheer quantity.
This contrast set is made explicit in discriminative or issue-sensitive captioning tasks \cite{ou2023pragmatic,cohn2018pragmatically,nie-etal-2020-pragmatic,andreas2016reasoning}, but is absent from common image description datasets \cite{ilinykh2018task,ilinykh2019tell,pezzelle2023dealing, takmaz2022less}. The idea also has precedent in the denotation graph of \citet{young-etal-2014-image}, where caption entailment and similarity relationships are defined by the sets of images a pair of descriptions truthfully applies to. We adopt the contrast-set approach to operationalize specificity using image-text compatibility scores: a description's specificity is determined by how well it distinguishes the target image from an \emph{implicit} set of alternatives.

\paragraph{Limitations of evaluation metrics} 

Existing evaluation metrics for image captioning fail to disentangle length from specificity. 
Reference-based metrics like BLEU \cite{10.3115/1073083.1073135}, ROUGE \cite{lin-2004-rouge}, and METEOR \cite{banerjee-lavie-2005-meteor} primarily assess similarity to human-written references but fail to capture human judgments of distinction \cite{kapur-kreiss-2024-reference}.
Referenceless metrics such as CLIPScore \cite{hessel2022clipscorereferencefreeevaluationmetric} measure image-text alignment but do not explicitly account for the contrastive value of the information provided \cite{kreiss-etal-2022-context}. 
None of these metrics capture specificity independent of length in communication-theoretic terms \cite{newman2020communication,tang2024grounding,coppock2020informativity}.
By constructing a dataset that manipulates length and information content independently, we provide a framework for evaluating specificity directly.

\section{Approach}

\subsection{Dataset construction}

To systematically investigate the relationship between length and specificity, we sampled 5,000 images uniformly across MS COCO’s 80 categories \cite{lin2015microsoftcococommonobjects}.
Our core theoretical contrast draws on possible-world semantics: descriptions expressing the same propositions should have equivalent specificity regardless of length, as they rule out the same possible images. 
Descriptions that incorporate distinct informational content should rule out more alternatives, yielding higher specificity. 
A metric that successfully disentangles length from specificity should detect this difference.
For each image, we generated multiple description variants that deliberately vary in length and content (see \autoref{tab:example_captions}):

\begin{description}[style=unboxed,leftmargin=0pt,itemsep=0pt,topsep=3pt]
    \item[\emph{Original}:] A single human-written description for the image from MS COCO.\vspace{-3pt}
    \item[\emph{Verbose}:] A longer rephrasing of the original that preserves the same semantic content, increasing length without adding new information.\vspace{-3pt}
    \item[\emph{Composite}:] A longer description that combines content from all five COCO reference descriptions, incorporating additional distinct details.\vspace{-3pt}
    \item[\emph{Image-to-Text}:] A VLM-generated description based on the image and minimal instructions.\vspace{-3pt}
\end{description}

The latter three description types were generated using OpenAI's \texttt{GPT-4o-mini} \cite[][prompts in \autoref{sec:appendix}]{hurst2024gpt}.
While the verbose and composite conditions provide a theoretical frame for analysis, the image-to-text condition provides a practical baseline for how VLMs balance length and specificity in practice. We make our complete dataset, experiments, and analyses available.\footnote{\url{https://github.com/rkapur102/vision-language-specificity}}

\subsection{Measuring specificity}
The central challenge of measuring specificity is that it is not defined on an absolute scale \cite{nie-etal-2020-pragmatic,degen2019redundancyusefulbayesianapproach}. 
Following the possible-worlds framework above, specificity must be defined relative to a contrast set: a description is more specific to the extent that it picks out the target from a set of implicit alternatives. 
For each image-description pair, we define the contrast set as the remaining 4,999 images. 
We then quantify specificity as a description's ability to discriminate the target image from these competitor images.
The intuition, grounded in entailment relationships \cite[see, e.g.,][]{montague1970universal,urquhart1973semantics}, is that more specific descriptions apply more selectively to a single image: e.g., all images showing ``an albacore'' show ``a fish'' but not vice versa.

To investigate this, we operationalize this idea using CLIPScore \cite{hessel2022clipscorereferencefreeevaluationmetric}. 
Its contrastive training objective makes it well-suited for measuring image-text compatibility in a discriminative setting \cite{ou2023pragmatic,takmaz2022less}. 
For each description, we compute its CLIPScore against the target image and all 4,999 alternatives. The rank of the target image is our specificity measure, where lower ranks indicate higher specificity (i.e., the description is less compatible with competitor images and more uniquely picks out the target).
See \autoref{sec:computational} for technical details.

\section{Results}

\subsection{Validating human specificity preferences} 
\label{human}

\begin{figure}[t]
    \centering
    \includegraphics[trim=0 2.1cm 0 0, clip, width=.49\textwidth]{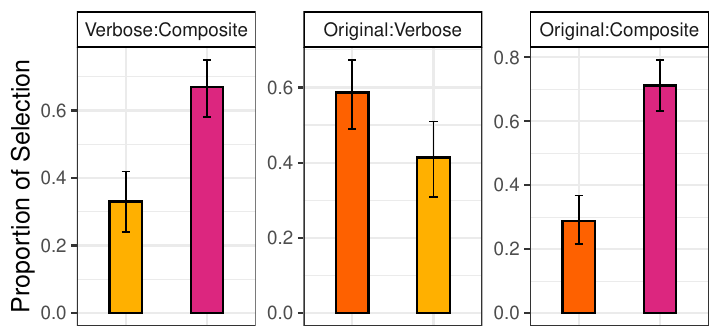}
    \caption{Pairwise human preferences by description type (95\% bootstrapped CIs). Full results in \autoref{pref_details}.}
    \label{fig:preference}
    \vspace{-10pt}
\end{figure}

With conditions and the metric established, we validated that human preferences track specificity over length.
We recruited 30 participants on Prolific. Participants saw an image paired with two descriptions and selected which they preferred. 
To isolate specificity, we sampled stimuli where verbose and composite descriptions were matched for length (see \autoref{pref_details}). Using logistic regression with length as a predictor, we found participants preferred composite descriptions over the original ($\beta = 1.85$, $z=2.54$, $p = .01$) and verbose descriptions ($\beta = -1.40$, $z=-4.6$, $p < .001$; see \autoref{fig:preference}).

To rule out potential confounds, we included average word frequency (excluding stopwords) as an additional fixed effect in our statistical models. Because formality and intended purpose are difficult to operationalize directly, we also included Flesch-Kincaid readability scores as a proxy, following prior work linking readability scores to text formality \cite{graesser2014coh}. Neither feature significantly predicts preferences ($p \in [0.25, 0.94]$ across models), whether incorporated individually or jointly, and all across-condition effects remain significant. 
These results confirm that the observed preferences are not explained by surface-level differences in word frequency or readability.

\subsection{Validating the specificity metric}
\label{specificity}

\begin{figure}[t]
    \centering
    \includegraphics[width=.85\linewidth]{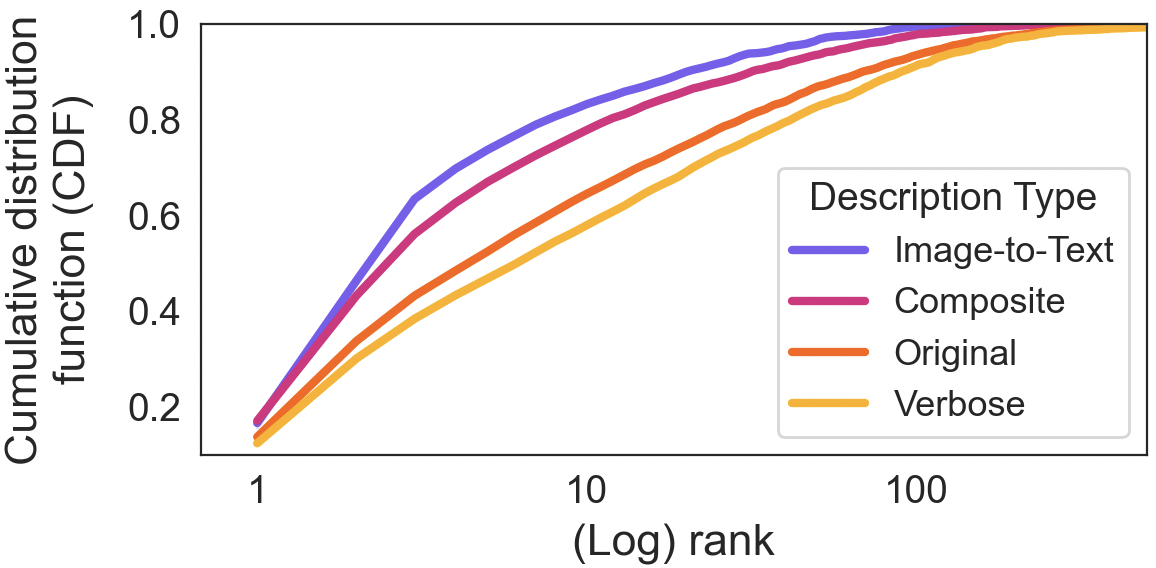}
    \caption{Cumulative distribution of ranks across description types relative to all other images.} 
    \label{fig:cdf}
    \vspace{-10pt}
\end{figure}

Having established that humans prefer more specific descriptions (not simply longer ones), we next test whether our specificity measure captures the same distinctions.
\autoref{fig:cdf} shows the cumulative distribution of target image ranks by description types. A steeper initial slope indicates that the descriptions more often receive low ranks (i.e., the target ranks highly), suggesting greater specificity.
The metric recovers a clear specificity hierarchy consistent with human preferences: composite descriptions yield significantly lower ranks than original ($\beta = -14.61$, $z=-10.33$, $p < 0.001$) and verbose ($\beta = -21.96$, $z=-13.72$, $p < 0.001$) descriptions. 
The difference between verbose and original ranks is also significant ($\beta=-7.35$, $z = -4.08$, $p < 0.001$), i.e., the verbose paraphrase appears to slightly decrease specificity rather than leaving it unchanged. 
Crucially, however, this does not affect the key comparison between length-matched composite and verbose conditions. 
We discuss a potential explanation related to CLIP's training data in the \hyperref[sec:limitations]{Limitations}.

\subsubsection{Testing hallucination robustness}
\label{hallucinations}

A natural concern is that hallucinated details could inflate specificity scores if they happen to be discriminative. To test this, we selected 37 composite captions manually verified to contain no hallucinations and used GPT-4o-mini to introduce targeted hallucinations, prompting: ``Change one detail in this description so that it becomes incompatible with the original. Only output the changed description and nothing else.'' We generated three hallucinated variants per caption and computed the CLIPScore of each against its matched image. 

Introducing hallucinations significantly worsened the specificity ranks ($t(36) = 2.50$, $p = 0.017$) with mean rank increasing from 17.05 (original composite) to 30.91 (average across three hallucinated variants per image). The drop in image-text alignment (as assigned by CLIPScore) outweighs any discriminative value gained from the hallucinated detail. That said, this measure is not a substitute for hallucination detection: it will be most reliable when hallucinations are minimal. We discuss this interaction further in the \hyperref[sec:limitations]{Limitations}.

\subsubsection{Analyzing contrast set sensitivity}
\label{contrast}

While we conducted our analysis on a large contrast set of 4,999 images to obtain high-resolution results, this may be too computationally expensive in practice. To investigate the sensitivity of our results to the size of the contrast set, we recomputed specificity ranks for all captions across the four main conditions (Image-to-Text, Composite, Original, Verbose) using subsampled contrast sets of $N={3, 10, 25, 50, 100, 250, 500, 1000, 2500, 4999[full]}$. Each subsampling was repeated 100 times per caption, and \autoref{tab:contrast_set_sizes} reports the mean rank across all captions per condition.

\begin{figure}[t]
    \centering
    \includegraphics[width=1.0\linewidth]{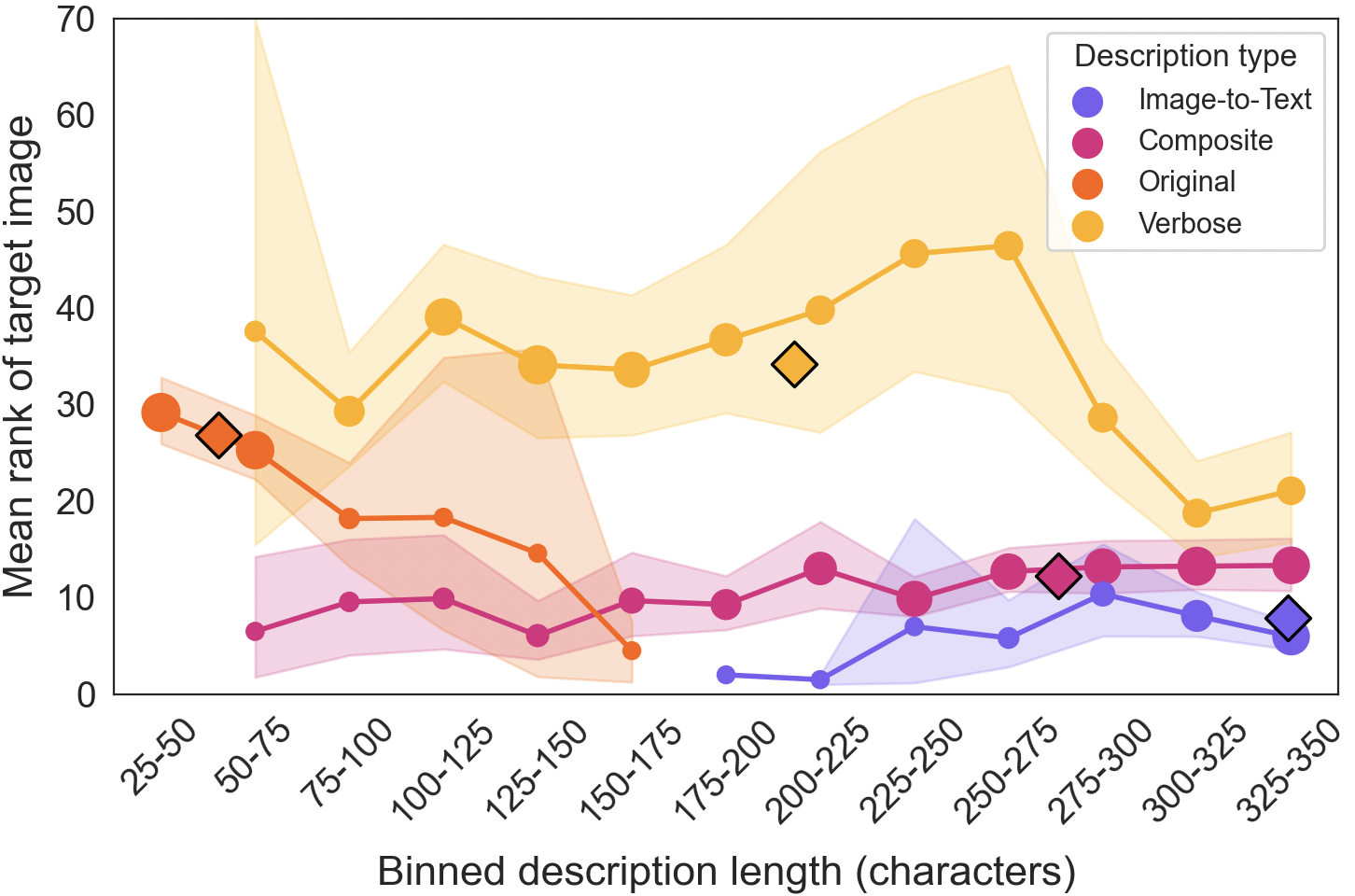}
    \caption{Mean rank vs. description length by type. Point size: bin sample size; ribbons: 95\% CIs; diamonds: overall means. Range excludes outliers.} 
    \label{fig:mean_rank_main}
    \vspace{-10pt}
\end{figure}

\begin{table*}[t]
\centering
\small
\begin{tabular}{lrrrrrrrrrr}
\hline
\textbf{Contrast set sizes} & \textbf{3} & \textbf{10} & \textbf{25} & \textbf{50} & \textbf{100} & \textbf{250} & \textbf{500} & \textbf{1000} & \textbf{2500} & \textbf{4999} \\
\hline
Image-to-Text & 1.004 & 1.014 & 1.034 & 1.069 & 1.139 & 1.345 & 1.693 & 2.391 & 4.456 & 7.911 \\
Composite     & 1.007 & 1.022 & 1.056 & 1.112 & 1.224 & 1.563 & 2.123 & 3.247 & 6.620 & 12.233 \\
Original      & 1.016 & 1.051 & 1.129 & 1.259 & 1.516 & 2.293 & 3.583 & 6.169 & 13.917 & 26.842 \\
Verbose       & 1.020 & 1.066 & 1.165 & 1.332 & 1.665 & 2.659 & 4.324 & 7.640 & 17.595 & 34.192 \\
\hline
\end{tabular}
\caption{Mean rank by description type across contrast set sizes $N$.}
\label{tab:contrast_set_sizes}
\vspace{-1em}
\end{table*}

We find that all pairwise between-condition differences are statistically significant (all $p < 0.001$) at every contrast set size tested and the ordering of conditions is preserved throughout (Image-to-Text, then Composite, then Original, then Verbose). This suggests that when aggregated across a dataset, the relative specificity differences are robust to the size of the contrast set. Larger contrast sets simply provide greater resolution in effect size magnitude. For example, Verbose vs. Image-to-Text yields $\beta=0.016$ at $N=3$ and $\beta=26.28$ at $N=4,999$, but both are highly significant. 
For application in open-ended settings, this suggests that a contrast set could be sampled or constructed from a large corpus like LAION-5B \cite{schuhmann2022laion5bopenlargescaledataset} in a manner that trades off contrast set size against computational resources, making these experiments more sustainable for subsequent research without sacrificing the robustness of relative specificity differences.

\subsection{Distinguishing specificity from length}

The preceding analyses are agnostic to the length of the descriptions. 
We now ask directly: does controlling for length eliminate the specificity differences between conditions? 
\autoref{fig:mean_rank_main} shows the mean rank as a function of description length for each condition, indicating that the specificity hierarchy persists across length bins.
In a regression model controlling for length as a covariate, composite descriptions remain more specific than original ($\beta = -16.97$, $z=-4.60$, $p < 0.001$, $\Delta R^2 = 0.002$) and verbose ($\beta = -21.35$, $z=-12.24$, $p < 0.001$, $\Delta R^2 = 0.018$) variants.

Beyond these differences, the within-condition relationship between length and specificity is itself revealing. 
Only original (human-written) descriptions show the expected relationship where longer descriptions are more specific ($\beta = -0.22$, $z=-2.42$, $p < 0.05$), suggesting humans genuinely add information with length. 
The other conditions lack this trend (verbose, image-to-text) or even show a reversal (composite: $\beta = 0.02$, $z=2.86$, $p < 0.01$). 
These patterns underscore that the ``longer means more specific'' heuristic cannot be assumed across data sources, particularly for synthetic or VLM-generated descriptions.
Finally, without specificity or length constraints, GPT-4o-mini produces descriptions significantly longer ($\beta = 60.97$, $z=26.56$, $p < 0.001$) and more specific ($\beta = -4.32$, $z=-4.15$, $p < 0.001$) than even composite descriptions. Composite descriptions are bound by what \textit{annotators} chose to write; VLM-generated captions reflect model choice in description, which shapes this difference. Evaluating how length constraints shape that choice is a natural next step toward understanding this interesting facet of model behavior.

\subsection{Evaluating VLM length constraints}
\label{ablations}

Having established that our approach distinguishes specificity from length, we can now investigate questions that the length-as-a-proxy assumption precluded. 
In particular, when we prompt VLMs to constrain their output length, does specificity decrease proportionally, or does it depend on how the constraint is imposed?
As a case study, we tested GPT-4o-mini under three length-constraint instructions:\vspace{-2pt}

\begin{description}[style=unboxed,leftmargin=0pt,itemsep=0pt,topsep=3pt]
    \item[\emph{Concise}:] Be as concise as possible.\vspace{-2pt}
    \item[\emph{Hard Constraint}:] Do not exceed 200 characters.\vspace{-2pt}
    \item[\emph{$k$-Limited}:] Do not exceed $k$ (i.e., the mean COCO caption length for that image).\vspace{-2pt}
\end{description}
\noindent The $k$-limited condition accounts for per-image variation in content, rather than imposing a uniform length across images. Full prompts are in \autoref{sec:appendix}.

All conditions significantly reduced description length, but their impact on specificity varied (\autoref{fig:mean_rank_ablation}). 
Counterintuitively, the concise condition is not associated with decreased specificity; it actually increased it ($\beta = -1.97$, $z=-3.12$, $p < 0.01$), suggesting that prompting for conciseness encourages the model to prioritize discriminative information.
Constraint strategy dictates specificity even at \textit{matched} lengths. 
The hard 200-character-limited descriptions are significantly less specific than concise ones ($\beta = 10.19$, $z=9.49$, $p < 0.001$, $\Delta R^2 = 0.013$). 
This indicates allocation matters: explicit length caps may lead to arbitrary truncation, while conciseness prompts allow the model to select what information to prioritize.

Notably, even the $k$-limited condition, despite being calibrated to image-specific COCO description lengths, does not replicate this human pattern; if anything, the VLM conditions show flat or reversed relationships between length and specificity. 
This echoes our earlier finding about the composite condition and underscores that matching length targets alone is insufficient to align VLM behavior with human patterns.
Together, these results demonstrate the practical value of measuring specificity independent of length: they reveal that prompt design choices have downstream consequences for specificity that length metrics alone would miss.

\section{Conclusion}

As VLMs become increasingly critical for making visual content accessible through image descriptions, we show that description length is not a reliable proxy for specificity, even though the two are frequently conflated. 
Using a contrast-set approach, we demonstrate descriptions can be lengthy yet vacuous, or concise yet dense. These differences matter for both human preferences and automated evaluation. 
Our findings call for evaluation metrics that measure specificity directly rather than relying on length as a surrogate, and for prompt design strategies that directly optimize for appropriate levels of specificity and relevance to context.

\section*{Limitations}
\label{sec:limitations}

Specificity is only one dimension among many that may matter for description quality. 
A maximally discriminative description could simply list every visible object in exhaustive detail, which would be accurate and specific, but potentially unreadable and irrelevant. 
Our focus on specificity complements rather than replaces attention to fluency, coherence, and user needs.

Our approach has two key dependencies, each of which brings its respective limitations. 
First, we operationalize specificity using CLIPScore, whose contrastive training objective made it a promising candidate. 
However, CLIPScore has known biases and practical constraints. 
Prior work has shown CLIP exhibits concept association biases \cite{Yamada2023,ahmadi2024examination} and struggles with spatial relationships \cite{kamath-etal-2023-whats}, which may affect which descriptive details register as discriminative under our metric. Additionally, the CLIP training data likely contained MSCOCO images and captions, thereby slightly inflating rank scores of the original caption-image pairs. The rewritten verbose descriptions may therefore have taken a minimal hit in specificity scores since they were never seen during training. While this can explain why descriptions from the verbose condition may receive slightly higher ranks than originals, the main effect of composite descriptions outranking verbose and originals cannot (see \autoref{specificity}).
CLIPScore also has a 77-token input limit that required us to exclude longer descriptions. Our framework is not committed to CLIPScore specifically; any model providing image-text compatibility scores could be substituted, potentially offering different sensitivity profiles. Importantly, because we compare conditions (verbose vs. composite vs. original) using the same encoder, any encoder blindness is a constant across conditions (as per the standard “within-subjects” logic) and doesn’t explain the real differences in specificity we observe between conditions (averaged over lots of data). We confirmed this by running a linear mixed-effects model predicting rank from description type with image as a random intercept, finding that all pairwise condition differences remain significant (p < 0.001, across statistical models). While the blindspots are a clear limitation (especially when trying to draw item-level instead of dataset-level inferences) these results suggest that they don’t significantly alter our main findings.

Though research shows CLIPScore and its extensions have general utility in detecting hallucinations \cite{oh2026vision, petryk2024aloha}, specific hallucinations that happen to be discriminative are most likely to survive the filter. Our specificity analysis is therefore complementary to hallucination detection, not a substitute for it. In our experimental data, hallucinated content is minimal because conditions are derived from human-written ground-truth captions, and we confirm in \hyperref[hallucinations]{Section 4.2.1} that artificially introduced hallucinations significantly worsen specificity scores. In practical applications, however, hallucinated content can meaningfully interfere with specificity estimates. It is precisely because specificity and faithfulness can interact in this way that independent tools for each are valuable: characterizing when hallucinations are specific versus generic requires an independent measure of specificity of the kind we propose.

Second, our operationalization of specificity is fundamentally relative to an implicit contrast set. 
There is no ``view from nowhere.''
The COCO-based contrast set we constructed was well-suited for our purposes: with 5,000 images sampled uniformly across 80 categories, we observed differences between conditions at the aggregate level.
More generally, the sensitivity of this approach will depend on the contrast set size (see \autoref{contrast}) and composition.
If the set is too small and/or lacks coverage, ceiling effects emerge; if there are too many images of a particular type (e.g., a specific kind of bird), it becomes disproportionately sensitive to variation within that dimension relative to other dimensions. 

We did observe ceiling effects in some cases, suggesting that a larger set may obtain more sensitive estimates.
And our uniform sampling approach was intended to mitigate biases in feature overrepresentation, though we cannot rule it out entirely.
For example, if green walls happen to be rarer in COCO than dual-toilet bathrooms, ``a bathroom with a green wall'' would counterintuitively rank as more specific than ``a bathroom with two toilet seats,'' even though the latter would statistically rule out higher proportion of real-world bathrooms. 
Importantly, however, such biases should not be correlated with description type, and should wash out in the aggregate analyses.
This concern underscores that our approach was tailored for dataset-level analysis rather than individual caption scoring.
Generalizing to absolute judgments about individual descriptions would require more sophisticated approaches, such as synthesizing contrastive images along specific semantic dimensions.

Finally, our LLM-generated conditions (verbose, composite, and the VLM ablations) may introduce stylistic artifacts beyond the intended manipulation, for example, formal hedging language (e.g. ``in the current situation, it appears...'') and potential differences in word frequency or register.
Our key comparison between verbose and composite at matched lengths partially controls for this, as both are subject to similar LLM stylistic tendencies, as well as our inclusion of some of these confounds as fixed effects in our statistical modeling (see \autoref{human}). However, individual items likely vary in how cleanly they instantiate the intended manipulation. 

The verbose condition is a controlled manipulation rather than a naturalistic sample of VLM verbosity. This is a deliberate feature of our design: isolating the length-specificity confound requires a condition that adds length without adding information, which we can then compare against the composite condition and, subsequently, against naturalistic VLM outputs (\autoref{ablations}). More broadly, our contrast-based ranking targets specificity and is agnostic about relevance. Consider two descriptions: one stating ``in the background, there is a house with a red roof'' and another stating ``in the foreground, there is a person with a hat.'' Under our framework, either could be more specific depending on how many images in the contrast set match each description. A background detail, while potentially less relevant to a user's needs, may be highly discriminative — and would score higher than non-specific padding that merely rephrases existing content. This is by design: we aim to capture whether added content narrows the set of compatible images, regardless of whether that content is relevant to a particular user's goals. Modeling relevance is beyond the scope of this paper.

Taken together, there are many challenging design choices when designing an approach to disentangle length and specificity. 
Despite that, we show that when making informed decisions about these constraints, such analyses provide insights that \emph{length as a proxy} can never deliver.

\section*{Acknowledgments}

We thank Google's GiG program for supporting this research. We are also grateful to Stanford's Social Interaction Lab, and the Models and Linguistic Theory Reading Group for feedback on earlier versions of this work.

\bibliography{custom}

\appendix

\section{Prompts}
\label{sec:appendix}

In this appendix, we describe the prompts used to generate each description type.

\subsection{Verbose} 
\label{verbose}
To synthesize a description with the same level of specificity as the \textit{original} description while increasing the length, we passed the \textit{coco} description to \texttt{GPT-4o-mini} with the following prompt: 
\begin{quote}
    Given this description, generate one longer description that expresses the same information as in the original description but in a more verbose way. In other words, use more words but say the same thing as given. Do not augment the description with any emotional or made-up information. Only output the longer description and nothing else.
\end{quote} 

\subsection{Composite}
To increase the expected level of specificity above and beyond the \textit{verbose} and \textit{original} conditions, we synthesized a \textit{composite} description by passing in \textit{all five} of the original descriptions from COCO with the following prompt: 
\begin{quote}
    Given these 5 descriptions, generate one longer, final description that combines all information in the individual descriptions. Do not augment the description with any emotional or made-up information. Only output the longer description and nothing else.
\end{quote}

\subsection{Image-to-Text}
To obtain a baseline for how a VLM describes the image (without reference to a human description), we simply passed the target image to GPT-4o-mini with the following prompt: 
\begin{quote}
    Describe this image and don't introduce any emotional information. Just describe what's there.
\end{quote} 

\begin{figure}[t]
    \centering
    \includegraphics[width=0.99\linewidth]{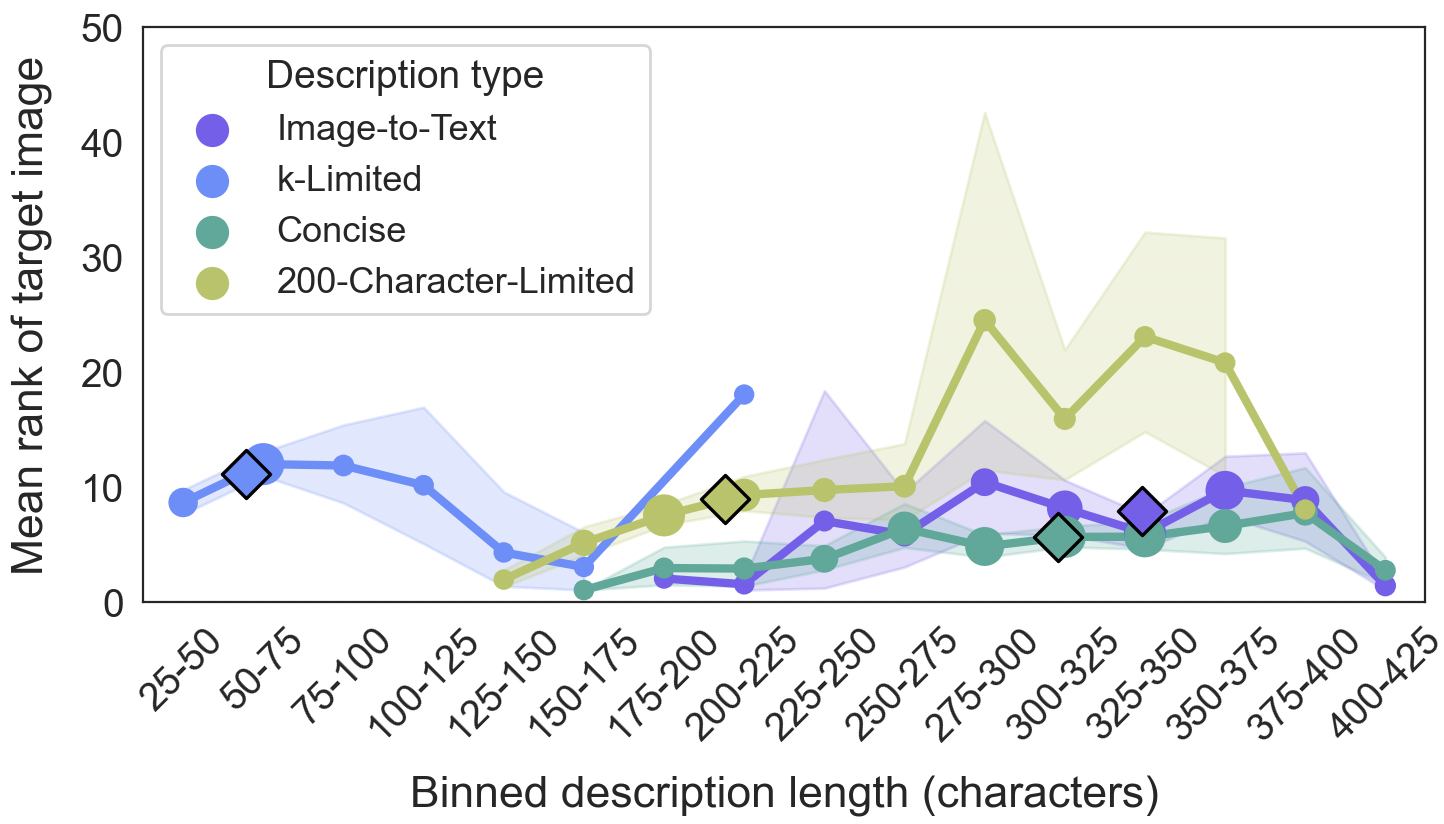}
    \caption{Mean rank vs. description length by each type in the length-constrained case study (\autoref{ablations}). Point size: bin sample size; ribbons: 95\% CIs; diamonds: overall means. Range excludes outliers.} 
    \label{fig:mean_rank_ablation}
\end{figure}

\subsection{Concise}
To understand how specificity changes when a VLM is instructed to constrain its length \textit{at its own discretion}, we simply passed the target image to GPT-4o-mini with the Image-to-Text prompt from above with the addition to "be as concise as possible": 
\begin{quote}
    Describe this image and don't introduce any emotional information. Just describe what's there. Be as concise as possible.
\end{quote} 

\begin{table*}[t]
\centering
\scriptsize
\begin{tabular}{|p{1.8cm}|p{6.8cm}|p{6.8cm}|}
\hline
\textsc{Image} & \includegraphics[width=\linewidth]{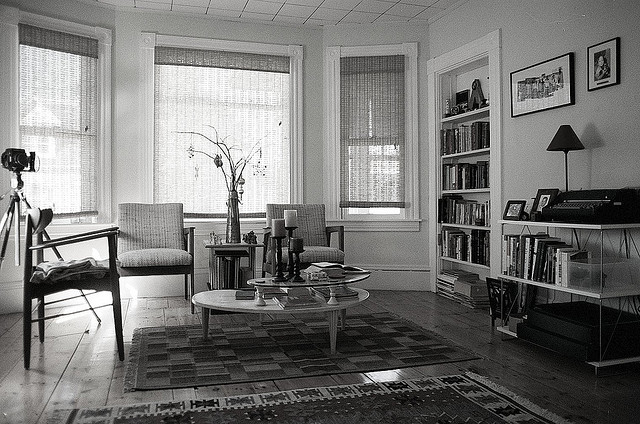} & \includegraphics[width=.88\linewidth]{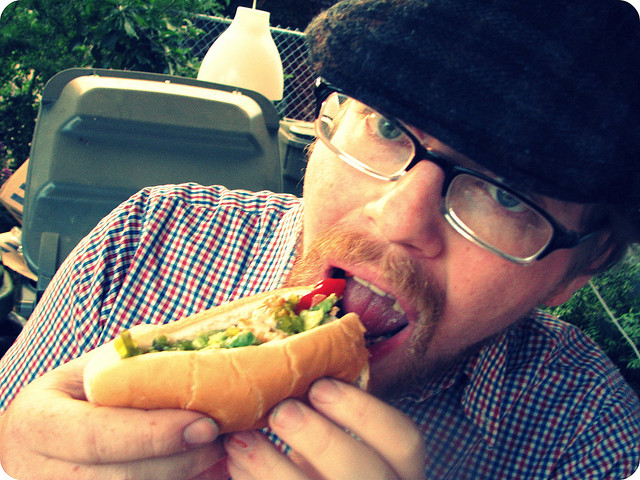} \\
  \hline
  \textsc{original} & A den with a table, chairs and a tripod camera. & A man wearing glasses while eating a hot dog. \\
  \hline
  \textsc{composite} & The den features a table and chairs, alongside a tripod camera. In the living room, there is a glass table, complemented by a black and white picture of a room that showcases three chairs and a coffee table. The living room itself includes a bay window, additional chairs, and a coffee table, creating a spacious area. The entire living area is very neatly organized, characterized by a black and white color scheme. & A man wearing glasses and a hat is taking a bite from a hot dog while dressed in a gingham shirt. The scene features a close-up of him eating the hot dog, highlighting his actions.  \\
  \hline
  \textsc{verbose} & The space referred to as a den is furnished with a table that serves as a central piece for various activities or gatherings, accompanied by several chairs that are arranged around it to provide seating. Additionally, this den is equipped with a tripod camera, which is a device typically used for stabilization while taking photographs or recording video, ensuring that it remains steady for better quality results. & A gentleman who is donning a pair of glasses is in the process of consuming a hot dog, which is a popular food item typically consisting of a cooked sausage served in a sliced bun. \\
  \hline
  \textsc{image-to-text} & The image shows a living room featuring natural light filtered through window blinds. There are three chairs arranged around a low, round coffee table at the center of the room. The table has several items on it, including a few books and candles. A small vase with plant stems is also present on the table. On one side, there is a bookshelf filled with books, and a lamp is positioned next to it. The floor is wooden, and there is a patterned rug beneath the coffee table. A tripod camera stands against the wall near the window. The walls are light-colored, and framed pictures are hung on one of the walls. The overall setting appears simple and organized. & The image shows a man holding a hot dog with various toppings, including green and red items, appearing to prepare to take a bite. He is wearing glasses and a dark cap and is dressed in a checkered shirt with a collar. In the background, there are some outdoor elements, including greenery and a container or object that looks like a storage bin. The lighting suggests it is a casual outdoor setting. \\
  \hline
  \textsc{concise} & The image depicts a living room with natural light filtering through window blinds. Three chairs are positioned around a low, round coffee table in the center of the room. The table holds several books, candles, and a small vase with plant stems. On one side, there is a bookshelf filled with books, accompanied by a lamp. The floor is wooden, and a patterned rug is placed beneath the coffee table. A tripod camera stands against the wall near the window. The walls are light-colored, with framed pictures hung on one of them. The setting appears simple and organized. & The image depicts a man holding a hot dog adorned with various toppings, including green and red ingredients. He is wearing glasses, a dark cap, and a checkered shirt with a collar. In the background, there is greenery and a storage bin or container, indicating an outdoor environment. The lighting suggests a casual setting. \\
  \hline
  \textsc{character-limited} & The living room features natural light through blinds, three chairs around a round coffee table with books, candles, and a vase. A bookshelf with a lamp, wooden floor, patterned rug, tripod camera, and framed pictures are present. & A man wearing glasses and a dark cap holds a hot dog with green and red toppings. He is dressed in a checkered shirt. The background features greenery and a storage bin. \\
  \hline

\end{tabular}

\caption{Example set of descriptions for images in our dataset. Composite and verbose descriptions are longer variants of the original description, but vary in the amount of additional information provided. \textsc{image-to-text}, \textsc{concise}, and \textsc{character-limited} are generated by a VLM (here, GPT-4o-mini) under different instruction conditions.}
\label{tab:example_captions2}
\end{table*}

\subsection{Character-limited}
Finally, to understand how specificity changes when a VLM is instructed to constrain its length \textit{to a hard cutoff}, we simply passed the target image to GPT-4o-mini with the Image-to-Text prompt from above and the additional instruction ``don't exceed 200 characters'' (example shown) or ``don't exceed \textit{k} characters'', where \textit{k} was passed in as the average COCO caption length for an image: 
\begin{quote}
    Describe this image and don't introduce any emotional information. Just describe what's there. Don't exceed 200 characters.
\end{quote} 

\begin{figure*}[t]
    \centering
    \includegraphics[width=1.0\linewidth]{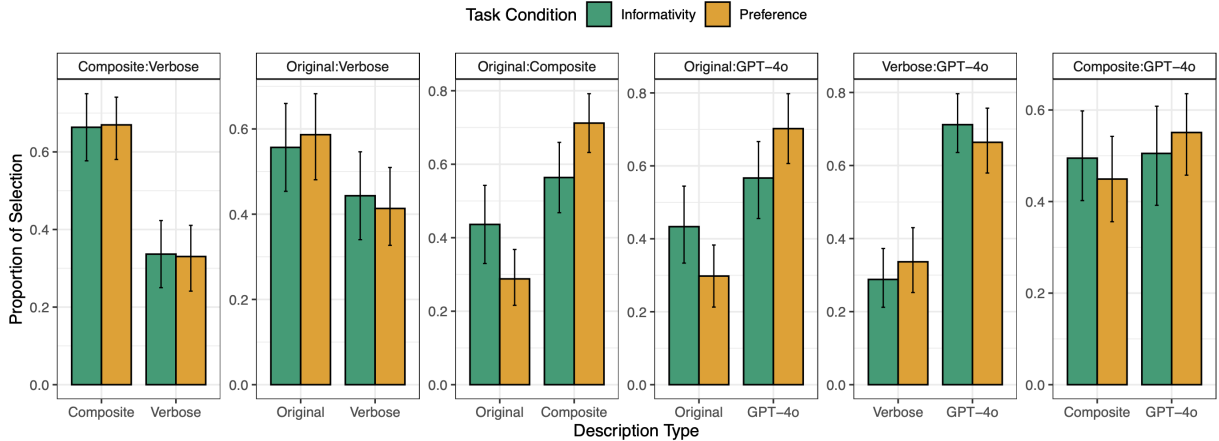}
    \caption{Experimental results of the preference study (in \autoref{human} and \autoref{app:expmain}) and the specificity study (in \autoref{app:expsuppl}).} 
    \label{fig:expall}
\end{figure*}

\section{Computational Details}
\label{sec:computational}

Computing specificity requires calculating CLIPScores between each description and all images in the contrast set. 
For our primary dataset (4 description types × 5,000 images × 5,000 contrast images), this amounts to 100 million pairwise scores. 
Using a single NVIDIA RTX 6000 Ada Generation GPU, this computation completed in approximately 5 hours. 
The additional length-constrained conditions (\autoref{ablations}) required an additional 2.5 hours.
While this particular size is feasible for research-scale evaluation, we have shown that our specificity metric is robust to varying contrast set sizes \hyperref[contrast]{(Section 4.2.2)}, which greatly improves scalability for larger deployments. There are several further optimizations to consider: parallelization across multiple GPUs, caching of image embeddings (which remain constant across descriptions), or sampling representative subsets from the contrast set rather than exhaustive comparison.
We also note that our approach is not committed to CLIPScore specifically; any model providing image-text compatibility scores could be substituted, potentially offering different tradeoffs between computational cost and sensitivity.

\section{Example Descriptions}
\label{descriptions}

We include 2 additional example sets of descriptions for images in our dataset, this time including the length-constrained ablations as well from \autoref{ablations}. These can be found in Table~\ref{tab:example_captions2}.

\section{Human experiment details}
\label{pref_details}

We conducted two human subject experiments to investigate people's understanding and preferences of the original and edited (i.e., composite and verbose) image descriptions from our dataset. \autoref{app:expmain} introduces the main study (reported in \autoref{human}) where we elicited people's preferences, and \autoref{app:expsuppl} supplements these findings with data on people's specificity judgments.

\subsection{Eliciting people's description preferences}\label{app:expmain}
Participants were recruited from the crowdsourcing platform Prolific, and recruitment was restricted to within the US, UK and Canada. Participants spent on average 10 minutes on the task and were paid \$14/hr. All data was anonymized before analysis. The anonymized data will be shared upon publication. The study was conducted under the lead author's institution's IRB protocol. The participant prompt for the preference study read as follows:

\begin{quote}
\small
\textbf{Thank you for participating in our study!}

In this study, you will see 30 images, each paired with two potential descriptions of the image. Your task is to determine which of the two descriptions you prefer. The whole study should take no longer than \textbf{10 minutes}.

Please do \textbf{not} participate on a mobile device, as the page may not display properly.

If you have any questions or concerns, please contact me at \textit{lead.author@email.address}

\vspace{-0.3em}
Please, enter your \textbf{Prolific ID}:
\end{quote}

\setlength{\parindent}{0pt}
After that, we displayed legal and IRB information for the participants to read. Then, once they clicked ``Begin Experiment,'' we displayed the following set of instructions: 

\begin{quote}
\small
In this study, you will see one image at a time, each paired with two potential descriptions.

Your task is to \textbf{choose the description that you would prefer to receive if you couldn't see the image}.

The descriptions you'll see vary in length and how much information they contain. Please note that \textbf{some descriptions might be long but still contain less information than shorter ones} and take that into account in your decision.

\end{quote}

We included the paragraph that highlighted the distinction between length and specificity due to the well-attested phenomenon that human raters in annotation studies often themselves use length as a heuristic for other measures in order to minimize cognitive load \cite{shen2023loose,malaviya2022cascading}.

\begin{figure}[t]
    \centering
    \fbox{
    \includegraphics[width=.99\linewidth]{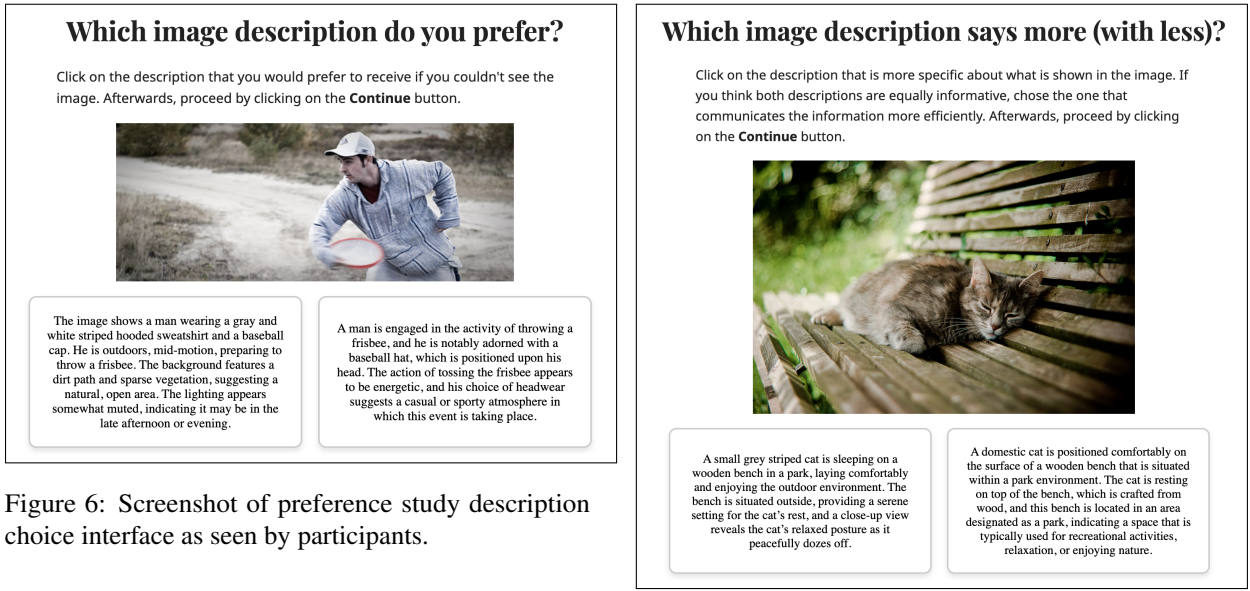}
    }
    \caption{Screenshot of preference study description choice interface as seen by participants.} 
    \label{fig:uxpref}
\end{figure}

\setlength{\parindent}{0pt}
Participants saw 30 images each with a pair of descriptions and chose the description they preferred (see a screenshot of this interface in \autoref{fig:uxpref}). 
Further results for the 3 VLM-generated description types can be found in \autoref{fig:mean_rank_ablation}.
With 30 participants each completing 30 trials, our design was powered for aggregate validation rather than detailed individual-differences analysis; characterizing sources of individual variation in description preferences remains a direction for future work.

\subsection{Eliciting people's specificity understanding}\label{app:expsuppl}

We conducted a second study that more directly tested whether the descriptions differed in their perceived specificity. As shown in \autoref{fig:uxinf}, the design is identical to the preference study, only the objective for participants changed. While participants in the first study were asked to select the description they \textit{preferred}, participants in this second study were asked to select the description that \textit{contains more information}. The only change to the main instructions was done to the second paragraph, which then read:

\begin{quote}
[...] Your task is to \textbf{choose the description that is more specific about the image content}. [...]
\end{quote}

\begin{figure}[t]
    \centering
    \fbox{
    \includegraphics[width=.99\linewidth]{figures/infstudy_screenshot.pdf}
    }
    \caption{Screenshot of specificity study description choice interface as seen by participants.} 
    \label{fig:uxinf}
\end{figure} 

\subsection{Results}

\autoref{fig:expall} presents the proportion of description selections from each study. All analyses were conducted using the \texttt{glm} function in \texttt{R} (\texttt{chosen $\sim$ condition + length}). Similar to the preference study, we find that participants consider the composite descriptions more informative than the original descriptions ($\beta=1.63, z(185)=2.4, p=0.02$) and their verbose counterparts ($\beta=-1.30, z(205)=-4.1, p<0.001$). These results clearly show that our data successfully manipulates perceived specificity.

Participant choices across studies are significantly correlated ($r= 0.21$, $p<0.001$), suggesting that participants across conditions preferred descriptions that are more specific. This is confirmed when we use the empirically elicited average specificity rate as a predictor for each item. The average specificity rate is a significant predictor for preference data in the verbose-composite ($\beta_{spec}=1.73, z(282)=3.95, p<0.001$) and original-composite settings ($\beta_{spec}=1.10, z(232)=2.34, p=0.02$), and marginally significant in the original-verbose setting ($\beta_{spec}=0.86, z(204)=1.90, p=0.057$).
These results show a strong association between the descriptions people prefer and how specific they perceive them to be, and that this goes beyond what can be explained through length.

\end{document}